\documentclass[11pt,a4paper]{article}
\usepackage[hyperref]{naaclhlt2019}
\usepackage{times}
\usepackage{latexsym}

\usepackage{url}
\usepackage{multirow}
\usepackage{graphicx}
\usepackage{setspace}
\usepackage{comment}

\usepackage{booktabs}
\usepackage{rotating}
\usepackage{multicol}
\usepackage{color}

\aclfinalcopy 
\def\aclpaperid{464} 

\title{WiC: the Word-in-Context Dataset \\ for Evaluating Context-Sensitive Meaning Representations}

 \author{
  Mohammad Taher Pilehvar$^{1,2}$ \and Jose Camacho-Collados$^{3}$ \\
  $^1$DTAL, University of Cambridge,  UK \\
  $^2$Tehran Institute for Advanced Studies (TeIAS), Tehran, Iran\\
  $^3$School of Computer Science and Informatics, Cardiff University,  UK\\
  {\tt mp792@cam.ac.uk, camachocolladosj@cardiff.ac.uk} \\
  }
\date{}

\begin{document}
\maketitle
\begin{abstract}
  By design, word embeddings are unable to model the dynamic nature of words' semantics, i.e., the property of words to correspond to potentially different meanings. 
  To address this limitation, dozens of specialized meaning representation techniques such as sense or contextualized embeddings have been proposed. However, despite the popularity of research on this topic, very few evaluation benchmarks exist that specifically focus on the dynamic semantics of words. In this paper we show that existing models have surpassed the performance ceiling of the standard evaluation dataset for the purpose, i.e., Stanford Contextual Word Similarity, and highlight its shortcomings. To address the lack of a suitable benchmark, we put forward a large-scale Word in Context dataset, called WiC, based on annotations curated by experts, for generic evaluation of context-sensitive representations. WiC is released in \url{https://pilehvar.github.io/wic/}.
\end{abstract}

\section{Introduction}

One of the main limitations of mainstream word embeddings lies in their static nature, i.e., a word is associated with the same embedding, independently from the context in which it appears.
Therefore, these embeddings are unable to reflect the dynamic nature of ambiguous words\footnote{Ambiguous words are important as they constitute the most frequent words in a natural language \cite{Zipf:49}.}, in that they can correspond to different (potentially unrelated) meanings depending on their usage in context \cite{camacho2018word}.
To get around this limitation dozens of proposals have been put forward, mainly in two categories: multi-prototype embeddings \cite{ReisingerMooney:2010,Neelakantanetal:2014,pelevina2016making}, which usually leverage context clustering in order to learn distinct representations for individual meanings of words, and contextualized word embeddings \cite{melamud2016context2vec,deepcontextual:2018}, which instead compute a single dynamic embedding for a given word which can adapt itself to arbitrary contexts for the word.

Despite the popularity of research on these specialised embeddings, very few benchmarks exist for their evaluation.
Most works in this domain either perform evaluations on word similarity datasets (in which words are presented in isolation; hence, they are not suitable for verifying the dynamic nature of word semantics) or carry out impact analysis in downstream NLP applications (usually, by taking word embeddings as baseline).
Despite providing a suitable means of verifying the effectiveness of the embeddings, the downstream evaluation cannot replace generic evaluations as it is difficult to isolate the impact of embeddings from many other factors involved, including the algorithmic configuration and parameter setting of the system.
To our knowledge, the Stanford Contextual Word Similarity (SCWS) dataset \cite{Huangetal:2012} is the only existing benchmark that specifically focuses on the dynamic nature of word semantics.\footnote{With a similar goal in mind but focused on hypernymy, \newcite{YogarshiHypernyms2017} developed a benchmark to assess the capability of automatic systems to detect hypernymy relations in context.} In Section \ref{sec:scws} we will explain the limitations of this dataset for the evaluation of recent work in the literature.

In this paper we propose WiC, a novel dataset that provides a high-quality benchmark for the evaluation of context-sensitive word embeddings.
WiC provides multiple interesting characteristics:
(1) it is suitable for evaluating a wide range of techniques, including contextualized word and sense representation and word sense disambiguation; (2) it is framed as a binary classification dataset, in which, unlike SCWS, identical words are paired with each other (in different contexts); hence, a context-insensitive word embedding model would perform similarly to a random baseline; and (3) it is constructed using high quality annotations curated by experts.

\begin{table}[t!]
\setlength{\tabcolsep}{4.0pt}

  \footnotesize
  \scalebox{1}{
  \begin{tabular}{c p{0.43\textwidth}}
    \toprule
    \bf F &
    There's a lot of trash on the \textit{bed} of the river |
    I keep a glass of water next to my \textit{bed} when I sleep \\
    \bf F & 
    \textit{Justify} the margins |
    The end \textit{justifies} the means    \\
    \bf T &
    \textit{Air} pollution |
    Open a window and let in some \textit{air} \\
    \bf T &
    The expanded \textit{window} will give us time to catch the thieves |
    You have a two-hour \textit{window} of clear weather to finish working on the lawn \\

    \bottomrule    
  \end{tabular}
  }
  \caption{Sample positive (T) and negative (F) pairs from the WiC dataset (target word in \textit{italics}).}
  \label{tab:examples}
\end{table} 

\section{WiC: the Word-in-Context dataset}

We frame the task as binary classification.
Each \textit{instance} in WiC has a target word $w$, either a verb or a noun, for which two contexts, $c_1$ and $c_2$, are provided.
Each of these contexts triggers a specific meaning of $w$.
The task is to identify if the occurrences of $w$ in $c_1$ and $c_2$ correspond to the same meaning or not.
Table \ref{tab:examples} lists some examples from the dataset.
In what follows in this section, we describe the construction procedure of the dataset.

\subsection{Construction}

Contextual sentences in WiC were extracted from example usages provided for words in three lexical resources:
(1) WordNet \cite{Fellbaum:98}, the standard English lexicographic resource; (2) VerbNet \cite{schuler2005verbnet}, the largest domain-independent verb-based resource; and (3) Wiktionary\footnote{\url{https://www.wiktionary.org/}}, a large collaborative-constructed online dictionary. 
We used WordNet as our core resource, exploiting BabelNet's mappings \cite{NavigliPonzetto:12aij} as a bridge between Wiktionary and VerbNet to WordNet. 
Lexicographer examples constitute a reliable base for the construction of the dataset, as they are curated in a way to be clearly distinguishable across different senses of a word.


\subsubsection{Compilation}

As explained above, the dataset is composed of instances, each of which contain a target word and two examples containing the target word. 
An instance can be either positive or negative, depending on whether the corresponding $c_1$ and $c_2$ are listed for the same sense of $w$ in the target resource.
In order to compile the dataset, we first obtained all the possible positive and negative instances from all resources, with the only condition of the surface word form occurring in both $c_1$ and $c_2$.\footnote{Given that WordNet provides examples for synsets (rather than word senses), a target word (sense) might not occur in all the examples of its corresponding synset.}
The total number of initial examples extracted from all resources at this stage were 23,949, 10,564 and 636 for WordNet, Wiktionary and VerbNet, respectively.
We first compiled the test and development sets with two constraints: (1) not having more than three instances for the same target word, and (2) not having repeated contextual sentences across instances. 
These constraints were enforced to have a diverse and balanced set which covers as many unique words as possible. With all these constraints in mind, we set apart 1,600 and 800 instances for the test and development sets, respectively. We ensured that all the splits were balanced for their positive and negative examples. 
The remaining 
instances whose examples did not overlap with test and development formed our initial training dataset. 
\paragraph{Semi-automatic check.}
Even though very few in number, all resources (even exprt-based ones) contain errors such as incorrect part-of-speech tags or ill-formed examples. Moreover, the extraction of examples and the mappings across resources were not always accurate. In order to have as few resource-specific and mapping errors as possible, all training, development and test sets were semi-automatically post-processed, either with small fixes whenever possible or by removing problematic instances otherwise.


\subsubsection{Pruning}
\label{pruning}

WordNet is known to be a fine-grained resource \cite{Navigli:06b}.
Often, different senses of the same word are hardly distinguishable from one another even for humans.
For example, more than 40 senses are listed for the verb \textit{run}, with many of them corresponding to similar concepts, e.g., ``move fast'', ``travel rapidly'', and ``run with the ball''. 
In order to avoid this high-granularity, we performed an automatic pruning of the resource, removing instances with subtle sense distinctions. 
Sense clustering is not a very well-defined problem \cite{mccarthy2016word} and there are different strategies to perform this sense distinction \cite{Snowetal:2007,Pilehvaretal:2013,mancini2017sw2v}. 
We adopted a simple strategy and removed all pairs whose senses were first degree connections in the WordNet semantic graph, including sister senses, and those which belonged to the same supersense, i.e. sense clusters from the Wordnet lexicographer files\footnote{\url{wordnet.princeton.edu/documentation/lexnames5wn}}. There are a total of 44 supersenses in WordNet, comprising semantic categories such as \textit{shape}, \textit{substance} or \textit{event}. 
This coarsening of the WordNet sense inventory has been shown particularly useful in downstream applications \cite{rud-EtAl:2011:ACL-HLT2011,severyn-nicosia-moschitti:2013:Short,flekovasupersense,pilehvaracl17}. 
In the next section we show that the pruning resulted in a significant boost in the clarity of the dataset.

\subsection{Quality check}
\label{sec:qcheck}

To verify the quality and the difficulty of the dataset and to estimate the human-level performance upperbound, we randomly sampled four sets of 100 instances from the test set, with an overlap of 50 instances between two of the annotators. Each set was assigned to an annotator who was asked to label each instance based on whether they thought the two occurrences of the word referred to the same meaning or not.\footnote{Annotators were not lexicographers. To make the task more understandable, they were asked if in their opinion the two words would belong to the same dictionary entry or not.}
The annotators were not provided with knowledge from any external lexical resource (such as WordNet). Specifically, the number of senses and the sense distinctions of the word (in the target sense inventory) were unknown to the annotators.

We found the average human accuracy on the dataset to be 80.0\% (individual scores of 79\%, 79\%, 80\% and 82\%). We take this as an estimation of the human-level performance upperbound of the dataset. For the overlapping section, we computed the agreement between the two annotators to be 80\%. 
Note that the annotators were not provided with sense distinctions to resemble the more difficult scenario for unsupervised models (which do not benefit from sense-based knowledge resources).
Having access to sense definitions/distinctions would have substantially raised the performance bar.

\paragraph{Impact of pruning.}
To check the effectiveness of our pruning strategy, we also sampled a set of 100 instances from the batch of instances that were pruned from the dataset. Similarly, the annotators were asked to independently label instances in the set. We computed the average accuracy on this set to be 57\% (56\% and 58\%), which is substantially lower than that for the final pruned set (i.e. 80\%). This indicates the success of our pruning strategy in improving the semantic clarity of the dataset.

\subsection{Statistics}

Table \ref{tab:statsdataset} shows the statistics of the different splits of WiC. 
The test set contains a large number of unique target words 
(1,256), reflecting the variety of the dataset. 
The large training split of 5,428 
instances makes the dataset suitable for various supervised algorithms, including deep learning models.
Only 36\% of the target words in the test split overlap with those in the training, with no overlap of contextual sentences across the splits. 
This makes WiC extremely challenging for systems that heavily rely on pattern matching.

\begin{table}
\setlength{\tabcolsep}{5.0pt}
\scalebox{0.84}{ 
\begin{tabular}{lcccc}
\toprule
\bf Split  &
\bf Instances &
\bf Nouns &
\bf Verbs &
\bf Unique words 
\\
\midrule
 Training & 5,428   & 49\% & 51\%  & 1,256       \\

  Dev & ~~~638 & 62\%  & 38\%   & ~~~599    \\
 Test &  1,400  & 59\%  & 41\%    & 1,184       \\
\bottomrule
\end{tabular}
}
\caption{\label{tab:statsdataset} Statistics of different splits of WiC.}
\end{table}

\section{Experiments}
\label{sec:experiments}

We experimented with recent multi-prototype and contextualized word embedding techniques. Evaluation of other embedding models as well as word sense disambiguation systems is left for future work.

\paragraph{Contextualized word embeddings.}
One of the pioneering contextualized word embedding models is {\bf Context2Vec} \cite{melamud2016context2vec}, which computes the embedding for a word in context using a multi-layer perceptron which is built on top of a bidirectional LSTM \cite{hochreiter1997long} language model. We used the 600-$d$ UkWac pre-trained models\footnote{\url{https://github.com/orenmel/context2vec}}.
{\bf ELMo} \cite{deepcontextual:2018} is a character-based model which learns dynamic word embeddings that can change depending on the context. ELMo embeddings are essentially the internal states of a deep LSTM-based language model, pre-trained on a large text corpus.
We used the 1024-$d$ pre-trained models\footnote{\url{https://www.tensorflow.org/hub/modules/google/elmo/1}} 
for two configurations:
ELMo$_1$, the first LSTM hidden state, and ELMo$_3$, the weighted sum of the 3 layers of LSTM.
A more recent contextualized model is {\bf BERT} \cite{devlin2018bert}. The technique is built upon earlier contextual representations, including ELMo,
but differs in the fact that, unlike those models which are mainly unidirectional, BERT is bidirectional, i.e., it considers contexts on both sides of the target word during representation.
We experimented with two pre-trained BERT models: {\it base} (768 dimensions, 12 layer, 110M parameters) and {\it large} (1024 dimensions, 24 layer, 340M parameters).\footnote{\url{https://github.com/google-research/bert/blob/master/}}
Around 22\% of the pairs in the test set had at least one of their target words not covered by these models. For such out-of-vocabulary cases, we used BERT's default tokenizer to split the unknown word to subwords and computed its embedding as the centroid of the corresponding subwords' embeddings. 

\paragraph{Multi-prototype embeddings.}
We experiment with three recent techniques that release 300-$d$ pre-trained multi-prototype embeddings\footnote{\textit{Multi-prototype} embeddings are also referred to as \textit{sense} embeddings in the literature.}.
{\bf JBT}\footnote{\url{https://github.com/uhh-lt/sensegram}} \cite{pelevina2016making} induces different senses by clustering graphs constructed using word embeddings and computes embedding for each cluster (sense).
{\bf DeConf}\footnote{\url{https://pilehvar.github.io/deconf/}} \cite{PilehvarCollier:2016emnlp} exploits the knowledge encoded in WordNet. For each sense, it extracts from the resource the set of semantically related words, called sense biasing words, which are in turn used to compute the sense embedding.
{\bf SW2V}\footnote{\url{http://lcl.uniroma1.it/sw2v}} \cite{mancini2017sw2v} is an extension of Word2Vec \cite{Mikolovetal:2013} for jointly learning word and sense embeddings, producing a shared vector space of words and senses as a result.
For these three methods we follow the disambiguation strategy suggested by \newcite{pelevina2016making}: for each example we retrieve the closest sense embedding to the context vector, which is computed by averaging its contained words' embeddings.

\paragraph{Sentence-level baselines.}
We also report results for two baseline models which view the task as context (sentence) similarity.
The {\bf BoW} system views the sentence as a bag of words and computes a simple embedding as average of its words.
The system makes use of Word2vec \cite{mikolov2013distributed} 300-$d$ embeddings pre-trained on the Google News corpus
. 
{\bf Sentence LSTM} is another baseline, which differently from the other models, does not obtain explicit encoded representations of the target word or sentence.
The system has two LSTM layers with 50 units, one for each context side, which concatenates the outputs and passes that to a feedforward layer with 64 neurons, followed by a dropout layer at rate 0.5, and a final one-neuron output layer of sigmoid activation. \newline

We used two simple binary classifiers in our experiments on top of all comparison systems (except for the LSTM baseline).
{\bf MLP}: a simple dense network with 100 hidden neurons (ReLU activation), and one output neuron (sigmoid activation), tuned on the development set (batch size: 32;  optimizer: Adam; loss: binary crossentropy). Given the stochasticity of the network optimizer, we report average results for five runs ($\pm$ standard deviation).
{\bf Threshold}: a simple threshold-based classifier based on the cosine distance of the two input vectors, tuned with step size 0.02 on the development set.

\begin{table}
\setlength{\tabcolsep}{11.0pt}
\footnotesize
\resizebox{\columnwidth}{!}
{
\begin{tabular}{lcc}
\toprule
&  \bf MLP &  \bf Threshold \\


\midrule
 \multicolumn{3}{l}{\bf Contextualized word-based models} \\ 
 \midrule 
  Context2vec  & 57.9 $\pm$ 0.9  & 59.3   
  \\
 ElMo$_1$  & 56.4 $\pm$ 0.6  & 57.7    \\
 ElMo$_3$  & 57.2 $\pm$ 0.8  & 56.5     \\
 BERT$_{base}$
 & \textbf{60.2} $\pm$ 0.4 & 65.4 \\
 BERT$_{large}$
 & 57.4 $\pm$ 1.0 & \bf 65.5\\
 
 \midrule
 \multicolumn{3}{l}{\bf Multi-prototype models} \\ 
 \midrule 
 DeConf* & 52.4 $\pm$ 0.8 & 58.7 \\
 SW2V*   &   54.1 $\pm$ 0.5  &   58.1 \\
 JBT    &   54.1 $\pm$ 0.6  &   53.6 \\
 \midrule
 \multicolumn{3}{l}{\bf Sentence-level baselines} \\ 
 \midrule
   BoW    &  54.2 $\pm$ 1.3  & 58.7 \\
   Sentence LSTM & \multicolumn{2}{c}{53.1 $\pm$ 0.9} \\
\bottomrule
\end{tabular}
}
\caption{\label{tab:results} Accuracy \% performance of different models on the WiC dataset. The estimated (human-level) performance is 80.0 (cf. Section \ref{sec:qcheck}) and a random baseline would perform at 50.0.
Systems marked with * make use of external lexical resources.
}
\end{table}

\subsection{Results}
\label{results}

Table \ref{tab:results} shows the results on WiC.
In general, the dataset proves to be very difficult for all the techniques, with the best model, i.e., 
BERT$_{large}$, providing around 15.5\% absolute improvement over a random baseline. Among the two classifiers, the simple threshold-based strategy, which computes the cosine distance between the two encodings, proves to be more efficient than the MLP network which might not be suitable for this setting with relatively small training data. The $\sim$15\% absolute accuracy difference between human-level upperbound and state-of-the-art performance suggests, however, a challenging dataset and encourages future research in context-sensitive word embeddings to leverage WiC in their evaluations. 

Among the LSTM-based contextualized models, Context2vec, which does not include the embedding of the target word in its representation, proves more competitive than ELMo.
However, surprisingly, neither ELMo nor Context2vec are able to significantly improve over the simple sentence BoW baseline, which in turn outperforms the sentence LSTM baseline. 
This raises a question about the ability of these models in capturing fine-grained semantics of words in various contexts. Finally, as far as multi-prototype techniques are concerned, DeConf is the best performer. We note that DeConf indirectly benefits from sense-level information from WordNet encoded in its embeddings. The same applies to SW2V, which leverages knowledge from a significantly larger lexical resource, i.e., BabelNet.
%
%

\section{Related work}
\label{sec:scws}

The Stanford Contextual Word Similarity (SCWS) dataset \cite{Huangetal:2012} comprises 2003 word pairs and is analogous to standard word similarity datasets, such as RG-65 \cite{RG65:1965} and SimLex \cite{hill2015simlex}, in which the task is to automatically estimate the semantic similarity of word pairs. Ideally, the estimated similarity scores should have high correlation with those given by human annotators.
However, there is 
a fundamental difference between SCWS and other word similarity datasets: 
each word in SCWS is associated with a context which triggers a specific meaning of the word.
The unique property of the dataset makes it a suitable benchmark for multi-prototype and contextualized word embeddings.
However, in the following, we highlight some of the limitations of the dataset which hinder its suitability for evaluating existing techniques.

Inter-rater agreement (IRA) is widely accepted as a metric to assess the annotation quality of a dataset.
The metric reflects the homogeneity of ratings which is expected to be high for a well-defined task and a qualified set of annotators.
For each word pair in SCWS ten scores were obtained through crowdsourcing.
We computed the pairwise IRA to be 0.35 (in terms of Spearman $\rho$ correlation) which is a very low figure.
The mean IRA (between each annotator and the average of others), which can be taken as a human-level performance upperbound, is 0.52.
%
Moreover, most of the instances in SCWS have context pairs with different target words.\footnote{Only 8\% (12\% if ignoring PoS) of SCWS pairs are identical but their assigned scores (by average 6.8) are substantially higher than the dataset average of 3.6 on a [0,10] scale.}
This makes it possible to test context-independent models, which only considers word pairs in isolation, on the dataset.
Importantly, such a context-independent model can easily surpass the human-level performance upperbound.
For instance, we computed the performance of the Google News Word2vec pre-trained word embeddings \cite{mikolov2013distributed} on the dataset to be 0.65 ($\rho$), which is significantly higher than the optimistic IRA for the dataset. In fact, \newcite{dubossarskycoming} showed how the reported high performance of multi-prototype techniques in this dataset was not due to an accurate sense representation, but rather to a subsampling effect, which had not been controlled for in similarity datasets.
In contrast, a context-insensitive word embedding model would perform no better than a random baseline on our dataset.

\section{Conclusions}

In this paper we have presented a benchmark for evaluating context-sensitive word representations. 
The proposed dataset, WiC, is based on lexicographic examples, which constitute a reliable basis to validate different models in their ability to perceive and discern different meanings of words.
We tested some of the recent state-of-the-art contextualized and multi-prototype embedding models on our dataset. The considerable gap between the performance of these models and the human-level upperbound suggests ample room for future work on modeling the semantics of words in context.

\section*{Acknowledgments}

We would like to thank Luis Espinosa-Anke and Carla P\'erez-Almendros for their help with the manual evaluation and Kiamehr Rezaee for running the BERT experiments. 

\bibliography{naaclhlt2019}
\bibliographystyle{acl_natbib}

\end{document}